\documentclass[a4paper,fleqn]{cas-dc}

\usepackage[numbers]{natbib}
\usepackage{algorithm}
\usepackage{algorithmic}
\usepackage{graphicx}
\usepackage[caption=false,font=normalsize,labelfont=sf,textfont=sf]{subfig}
\usepackage[section]{placeins}
\usepackage{amsthm}

\makeatletter
\newenvironment{breakablealgorithm}{
  \begin{center}
  \refstepcounter{algorithm}% New algorithm
  \hrule height.8pt depth0pt% @fs@pre for @fs@ruled
  \renewcommand{\caption}[2][relax]{% Make a new caption
    \raggedright\textbf{\ALG@name~\thealgorithm}##2\par%
    \ifx##1\relax % is relax
      \addcontentsline{loa}{algorithm}{\protect\numberline{\thealgorithm}##2}%
    \else % is not relax
      \addcontentsline{loa}{algorithm}{\protect\numberline{\thealgorithm}##1}%
    \fi
    \kern2pt\hrule\kern2pt
  }
}{
  \kern2pt\hrule\relax% @fs@post for @fs@ruled
  \end{center}
}
\makeatother
\def\tsc#1{\csdef{#1}{\textsc{\lowercase{#1}}\xspace}}
\tsc{WGM}
\tsc{QE}
\newdefinition{rmk}{Remark}
\newproof{pf}{Proof}
\newproof{pot}{Proof of Theorem \ref{thm}}
\theoremstyle{definition} % 这两行是后加的

\begin{document}
\let\WriteBookmarks\relax
\def\floatpagepagefraction{1}
\def\textpagefraction{.001}

\shorttitle{}

\shortauthors{Sirui Li et al.}

\title [mode = title]{AutoHGNN: Robust and Efficient Neural Architecture Search for Hypergraph Neural Networks}

\affiliation[1]{organization={Faculty of Electrical Engineering and Computer Science},
    addressline={Ningbo University},
    city={Ningbo},
%     citysep={,},
    postcode={315211},
    state={Zhejiang},
    country={China}
}
\affiliation[2]{organization={Department of Computer Science and Technology},
    addressline={University of Cambridge},
    city={Cambridge},
%     citysep={,},
    postcode={CB2 1TN},
    state={Cambridgeshire},
    country={United Kingdom}
}
% \affiliation[3]{organization={Department of Computer Science},
%     addressline={City University of Hong Kong},
%     city={Kowloon Tong},
% %     citysep={,},
%     postcode={999077},
%     state={Hong Kong},
%     country={China}
% }

\author[1]{Sirui Li}%[<options>]

\ead{2411100282@nbu.edu.cn}

\credit{Conceptualization, Formal Analysis, Investigation, Methodology, Software, Writing - original draft}

\author[2]{Pietro Liò}
\ead{Pietro.Lio@cl.cam.ac.uk}
\credit{Formal analysis, Project administration, Resources, Supervision, Writing - review \& editing}

\author[1]{Xinsheng Li}
\ead{2311100292@nbu.edu.cn}
\credit{Methodology, Software, Writing - review \& editing}

% \author[3]{Weidun Xie}%[]
% % Footnote of the second author
% % \fnmark[2]
% % Email id of the second author
% \ead{weidunxie2-c@my.cityu.edu.hk}
% % URL of the second author
% % \ead[url]{}
% % Credit authorship
% \credit{Writing review - editing, Software}

% \author[3]{Ka-chun Wong}%[]
% % Email id of the second author
% \ead{kc.w@cityu.edu.hk}
% % URL of the second author
% % \ead[url]{}
% % Credit authorship\
% \credit{Methodology, Software, Writing - Review}

\author[1]{Baisong Liu}
\ead{liubaisong@nbu.edu.cn}
\credit{Software, Resources, Investigation, Writing - review \& editing}

\cortext[1]{Corresponding author}
\author[1]{Chengbin Peng}%[]
\cormark[1]

\ead{pengchengbin@nbu.edu.cn}

\credit{Conceptualization, Formal Analysis, Funding acquisition, Investigation, Resources, Supervision, Writing - original draft, Writing - review \& editing}


\begin{abstract}
Hypergraph neural networks have achieved significant success in recent years. However, manual architecture crafting is labor-intensive and often fails to capture complex, higher-order relations, making the automation of hypergraph neural network structure design crucial. To improve the automation and adaptability of hypergraph learning, this paper proposes AutoHGNN, a neural architecture search framework tailored for hypergraph neural networks. First, we introduce a Hyper-Interaction Module (HIM) into the search space to address the mismatch between conventional graph neural network designs and hypergraph data. Second, we propose Hypergraph Stable Topological Distance (HyperSTD) as a structural selection criterion to identify architectures that best preserve the intrinsic structural affinities of the original hypergraph during differentiable search. Extensive experiments on various benchmark datasets demonstrate that AutoHGNN consistently outperforms manually designed and automatically searched baselines in classification accuracy and time efficiency, proving that the discovered architectures are significantly more effective.
\end{abstract}


% \begin{highlights}
% \item Hypergraph neural networks are designed automatically with stablity and efficiency.
% \item Search space models vertex-hyperedge message-passing with feature aggregation.
% \item Topology-aware metric selects candidates to improve final model choice.
% \end{highlights}

%\nocite{*}
% Keywords

\maketitle
\section{Introduction}
Graph structured data is fundamental to machine learning in domains such as recommender systems, social networks, and biochemistry. Graph Neural Networks (GNNs) have proven effective in modeling the pairwise relationships within these structures \cite{suvery_on_graph,graph_recommend_kbs,graph_recommend_pr,graph_chemistry}. Many real-world systems involve higher-order interactions that are not captured by those simple edges. Hypergraphs generalize graphs by representing group-wise relations as hyperedges, enabling the modeling of group interactions. For example, in academic research, a group of authors co-writing a paper forms a hyperedge connecting all author nodes, capturing the collaborative relation beyond pairwise co-authorship. Hypergraph Neural Networks (HGNNs) \cite{survey_on_hypergraph, hgnn, DHCF, hypergcn, hnhn, hypersage, hgnnp,unignn,unig_encoder,hypergraph_heterophily,whatsnet} have been proposed to leverage this structure, showing advantages in tasks where group context matters.
The study of hypergraph learning can broadly be divided into two main categories, including spectral approaches \cite{hgnn, hypergcn, DHCF, hnhn} and spatial methodologies \cite{hypersage, hgnnp,unignn,unig_encoder,whatsnet,hypergraph_heterophily}. Spectral HGNNs involved complex Laplacian transformations \cite{hgnn,hypergcn} or converting hypergraphs into traditional graphs \cite{DHCF,hypergcn,hnhn} to leverage conventional graph techniques, but such transformations tend to discard some of the higher-order structural information and introduce additional noise into the process. {Spatial methods adopted a different approach} by defining multi-level message propagation directly on the incidence matrix. They are able to preserve the original semantics of group relationships. These approaches implement message aggregation for both vertices and hyperedges, employing methods like degree normalization \cite{hypersage,hgnnp}, hyperedge weighting \cite{hgnnp}, and linear projections \cite{hgnnp,unig_encoder} to enhance numerical stability and expressive power while significantly improving the time efficiency.

%引出问题，下面开始总结nas方面内容
Meanwhile, Neural Architecture Search (NAS) has been making serious headway in automatically designing GNNs in recent years \cite{gnas_survey}. Drawing inspiration from prior breakthroughs in vision and language \cite{nas_review}, researchers adapted both search spaces and optimizers to the graph domain, exploring reinforcement learning (RL) and evolutionary algorithm (EA) strategies \cite{graphnas,auto_gnas,deepgnas,genetic-gnn,knowledge-aware} as well as one-shot and differentiable search paradigms \cite{PAS,search_to_capture_long_range,am_gnas,one_shot_gnas,d2gnas}. Work along the search-space axis has shown that carefully tailored operators and multi-hop aggregation choices substantially improve downstream performance, such as graph-classification oriented spaces and designs that explicitly model deeper stacking or multi-hop aggregation \cite{PAS,search_to_capture_long_range,am_gnas}. Complementary efforts focus on search efficiency and stability, introducing supernet weight sharing, greedy pruning, and decoupled optimization schemes to reduce the cost and the architecture-weight coupling that plagues naive relaxations \cite{one_shot_gnas,d2gnas}.

However, existing neural architecture search methods are primarily developed for traditional graphs and exhibit fundamental limitations when applied to hypergraph neural networks. Specifically, their search spaces fail to model the intrinsic vertex-hyperedge message passing of hypergraphs, which often leads to unstable optimization and architecture selection that neglects higher-order topology.

To overcome these limitations, we propose AutoHGNN, a differentiable NAS framework tailored for hypergraph neural networks. AutoHGNN designs a search space that explicitly encodes hypergraph-specific message propagation, employs a differentiable search strategy for stable optimization, and introduces a stabilized topological distance as a topology-aware criterion to select architectures that preserve higher-order structural relationships.
To sum up, it is a novel work to integrate the differentiable NAS strategy with a topology-aware selection method to search for HGNN architectures with vertex-hyperedge message-passing and feature aggregation. Our main contributions can be concluded as follows:
\begin{itemize}
        \item We design a hypergraph search space that is built upon a Hyper-Interaction Module (HIM) backbone with feature aggregation, explicitly exposing the aggregation operators, which directly captures the higher-order relations in hypergraph data.
        \item We introduce the Hypergraph Stable Topological Distance (HyperSTD) metric, a topology-aware module for robust architecture selection by evaluating structural fidelity beyond mere validation accuracy after the differentiable search.
        \item Across multiple public datasets containing both traditional graphs and hypergraphs, AutoHGNN consistently outperforms baseline models in accuracy while demonstrating significantly improved time efficiency.
\end{itemize}
% To be more precise, the search space of AutoHGNN is built for the two-stage propagation paradigm in hypergraphs, systematically exposing  alongside feature aggregation options. This comprehensive operator collection encompasses multiple implementation variants, while the fusion module facilitates the extraction and integration of features from multi-layer encoders. The decoupled differentiable sampler adds Gumbel noise to the logits space and samples via softmax, followed by fixing the network weights on the validation partition and updating the architecture parameters.  After obtaining the pruned candidate set, we compute the HyperSTD for each candidate. This metric measures the similarity between the predicted label matrix and the true label matrix under the hypergraph connectivity operator. Candidates with the smallest distance values enter the short-term evaluation pool. The architecture with the highest validation score is ultimately selected as the final model.
\section{Related Work}
\begin{table*}[htbp]
\centering
{
\caption{Comparison of representative hypergraph neural network methodologies.}
\label{tab:hgnn_comparison}
\begin{tabular}{@{}llll@{}}
\toprule
Method & Type & Message Passing & Complexity \\
\midrule
HGNN~\cite{hgnn} & Spectral & Laplacian & $\mathcal{O}(n^2)$ \\
HyperGCN~\cite{hypergcn} & Spectral & Graph conversion & $\mathcal{O}(m \cdot d)$ \\
HNHN~\cite{hnhn} & Spectral & Hyperedge neurons & $\mathcal{O}(n \cdot d)$ \\
HyperSAGE~\cite{hypersage} & Spatial & Two-stage aggregation & $\mathcal{O}(\text{sampling})$ \\
HGNN+~\cite{hgnnp} & Spatial & Aggregation + update & $\mathcal{O}(n d + m d)$ \\
UniG-Encoder~\cite{unig_encoder} & Spatial & Projection-encoding-decoding & $\mathcal{O}(n d)$ \\
WHATsNET~\cite{whatsnet} & Spatial & Attention + positional encoding & $\mathcal{O}(n^2)$ \\
\textbf{AutoHGNN (ours)} & Spatial (NAS) & Searchable two-stage aggregation & Low (fixed after search) \\
\bottomrule
\end{tabular}
}
\end{table*}
\subsection{Hypergraph Neural Networks}
% 早年方法多采用Spectral形式，如hgnn, hypergcn, dhcf等
Hypergraph neural networks (HGNNs) extend graph neural networks to encompass more complex relationships within hypergraphs by enabling node interactions via the hyperedges, becoming a powerful tool for higher-order interactions \cite{survey_on_hypergraph}. Early spectral methods such as HGNN \cite{hgnn}, HyperGCN \cite{hypergcn}, HNHN \cite{hnhn} and DHCF \cite{DHCF} contain complex Laplacian operators, or extend the provided hypergraph to a graph and then adopt methods in standard GNNs, thereby discarding complex higher-order connections within the hyperedge.
So recent HGNN works have highlighted the great significance of message-passing mechanism among nodes and hyperedges \cite{hypersage,hgnnp,unig_encoder,whatsnet,hypergraph_heterophily}. HyperSAGE \cite{hypersage} is a pioneering architecture that redefines hypergraph representation learning through a dual-layer message-passing process consisting of intra-hyperedge and inter-hyperedge aggregation. It employs neighborhood sampling to balance expressiveness with computational efficiency. HGNN+ \cite{hgnnp} further introduces a process for hyperedge-group formation to construct a hypergraph from any given data, incorporating adaptive fusion and a clear two-step spatial convolution process from vertex to hyperedge and back to vertex, both steps have an aggregation and an update stage. UniG-Encoder \cite{unig_encoder} formulates graph and hypergraph representation learning as a principled projection-encoding-decoding message-passing framework that preserves the interpretability of classical message passing. WHATsNET \cite{whatsnet} proposes edge-dependent problem and does message-passing with attention and positional encoding strategies. Li et al. \cite{hypergraph_heterophily} highlight the importance of heterophily in hypergraph data, noting that most existing HGNNs implicitly assume homophily and thus can fail when hyperedges connect dissimilar nodes. {Table \ref{tab:hgnn_comparison} shows the comparison of different HGNN methodologies.}

However, currently the majority of HGNNs' architectures are crafted by hand, which requires a lot of domain knowledge. In this work, we propose a neural architecture search approach to adaptively discover the best message-passing mechanism and identify latent relationships for hypergraph data.

\subsection{Neural Architecture Search on Graphs}
Recent years have seen rapid growth in methods for automatically designing graph neural network architectures. Many efforts adapted NAS paradigms from computer vision and NLP, primarily reinforcement-learning and evolutionary strategies, to search over aggregation operators, layer depth, and readout designs \cite{graphnas,auto_gnas,deepgnas,GNAS++,autograph,genetic-gnn,knowledge-aware}. Because these approaches are computationally expensive on graphs, the field rapidly shifted toward differentiable graph NAS (DGNAS), which leverages continuous relaxations and weight sharing to improve efficiency. One line of DGNAS work reduces cost by designing task-aware or compact search spaces. For example, PAS \cite{PAS} extends the search space to pooling choices for graph classification, LRGNN \cite{search_to_capture_long_range} emphasizes depth by autonomously constructing deep stacked GNNs to capture long-range dependencies, and AM-GNAS \cite{am_gnas} explicitly decomposes each layer into multi-hop aggregation and feature-fusion phases to broaden the searchable operations beyond conventional 1-hop designs.
A complementary line of work improves the search strategy itself to lower expenses and improve robustness \cite{one_shot_gnas, d2gnas, auto_heg}. For example, DSS \cite{one_shot_gnas} introduces a greedy supernet pruning method for DGNAS. The core idea revolves around assessing whether pruning the supernet is a viable way to increase the manageability and efficiency of differentiable graph NAS. D2GNAS \cite{d2gnas} uses decoupled sampling and single-path supernet schemes to mitigate architecture-weight coupling bias and further scale differentiable search. Auto-HeG \cite{auto_heg} considers the heterophily of graphs. It aims to create GNN models that excel in learning across different types of nodes, and introduces a novel distance metric that improves selection robustness under heterophily. {Methods like HGNAS \cite{hgnas} and AMP-HGNAS \cite{adaptivemetapath} extend NAS methods to heterogeneous graphs.}

Since most existing graph NAS work targets traditional graphs with binary edges, automatic search for hypergraph-specific components remains underexplored. In this paper, we endeavor to utilize differentiable NAS for hypergraph neural network architectures.

\section{Proposed Method}
\begin{figure*}[htb!]
\centering
\includegraphics[width=140mm]{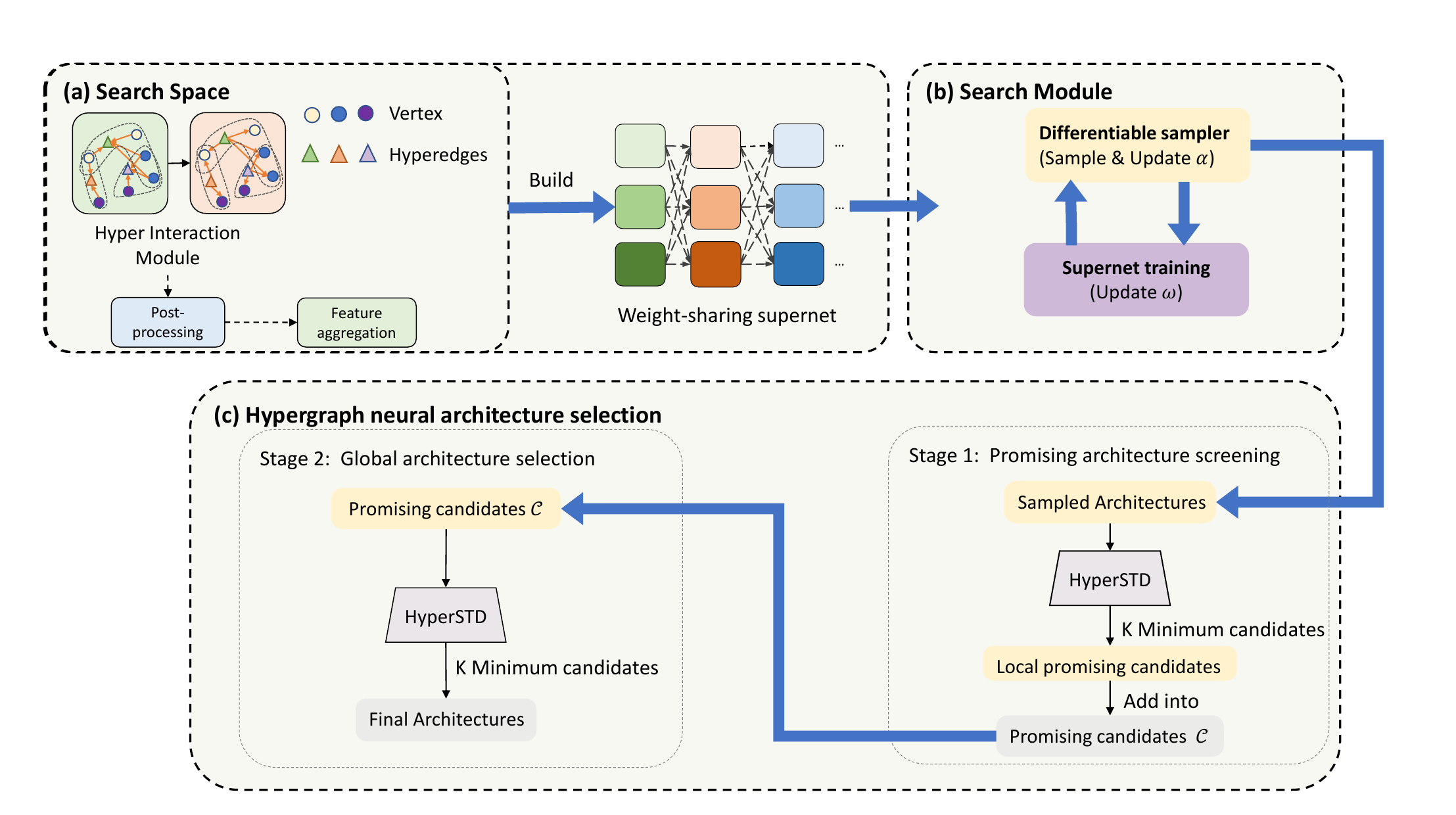}
\caption{The overview of AutoHGNN. Green, orange and blue denote vertex-to-hyperedge aggregation, hyperedge-to-vertex aggregation and post-processing operations respectively. Varying depth of same color indicates different members of the same operation type that share the same weight $\omega$. (a) The HGNN search space whose details are in Fig. \ref{fig:space}. (b) The search module with a weight-sharing supernet constructed from search space, a differentiable sampler and a supernet training modules. (c) Hypergraph neural architecture selection. In stage 1, promising candidates are selected from the j-th round of sampling and gradually merged into promising candidates set, so that the final architectures can be selected from them in stage 2.}
\label{fig:overview}
\end{figure*}

\subsection{Notations}
Let \emph{G} be a hypergraph with \emph{n} nodes which has $C$ different classes and \emph{m} hyperedges. $V=\{v_1, v_2, ..., v_n\}$ is the set of nodes and $E=\{e_1, e_2, ..., e_m\}$ is the set of hyperedges. We utilize $X_v^{(l)}\in\mathbb{R}^{n\times{d_V}} $ to represent node features and $X_e^{(l)}\in\mathbb{R}^{m\times{d_E}} $ to represent hyperedge features in network layer \emph{l}, where $d_V$ and $d_E$ denote the dimension of vertex and hyperedge features respectively. $H\in\{0,1\}^{n\times m}$ is the incidence matrix of \emph{G}, whose entries are defined as $H(v,e) = 
\begin{cases}
1, & \text{if } v \in e \\
0, & \text{if } v \notin e
\end{cases}.\, D_v\in \mathbb{R}^{n \times n}$ and $D_e\in \mathbb{R}^{m \times m}$ are vertex and hyperedge degree matrices.

\subsection{Search Space}
While a graph is a special format of hypergraph, the search space utilized in the graph NAS techniques can not be simply transferred to hypergraph NAS because they can not identify higher order relations in hypergraphs. Consequently, it's important to devise a search space that is tailored specifically for the intricacies of NAS for hypergraphs.

\begin{figure*}[htb!]
\centering
\includegraphics[width=140mm]{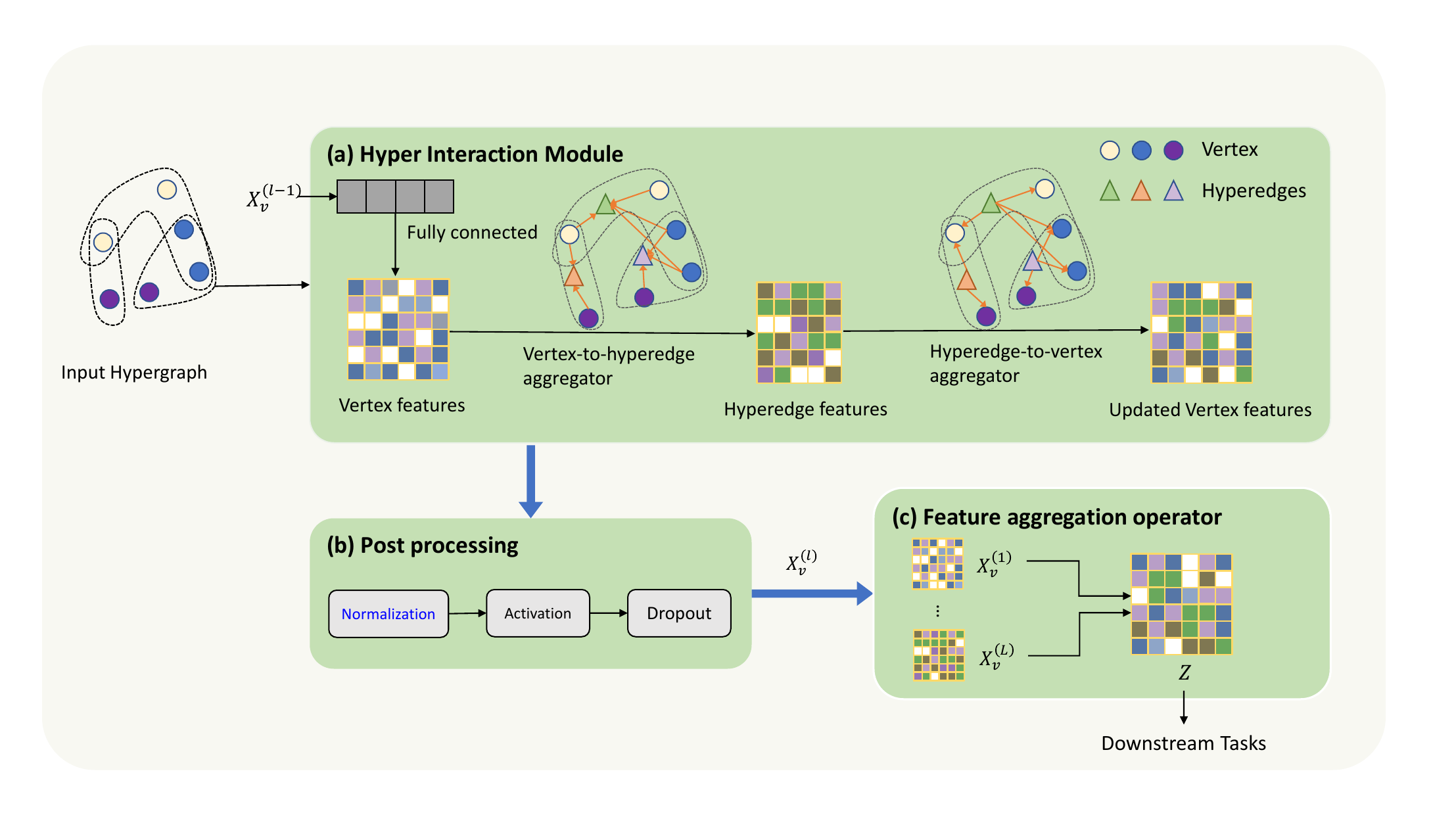}
\caption{\textbf{Definition of search space in AutoHGNN.} Each network layer consists of (a) a Hyper-Interaction Module layer with a vertex-to-hyperedge aggregator and a hyperedge-to-vertex aggregator and (b) a post-processing layer. All layers' outputs are fused by the (c) feature aggregation operator $\mathcal{F}$ to get final output $Z$ for downstream tasks. $X_v^{(l)}$ means vertex features of the $l$-th layer.}
\label{fig:space}
\end{figure*}

Our HGNN search space contains $L$ layers. As depicted in Fig. \ref{fig:space}, Each layer encompasses three core components, including a vertex-to-hyperedge aggregation layer, a hyperedge-to-vertex aggregation layer and a post-processing layer that includes normalization, activation and dropout operators. These layers serve as the encoder, transforming input hypergraph data into embeddings. A feature aggregation operator is employed to fuse their outputs.

\subsubsection{Hyper-Interaction Module}
% 这个篇幅要长点
The {vertex-to-hyperedge and hyperedge-to-vertex} aggregators form the core part of the space, namely the Hyper-Interaction Module mechanism. We implement the two-stage message-passing at every layer, which explicitly constructs and propagates high-order information through hyperedges before returning messages to nodes.

At Layer \emph{l} of the network we firstly use a fully connected layer to update original features or features learned in last layer, and transform it into target dimension,
\begin{equation}
        \tilde{X_v}^{(l-1)}=fc(X_v^{(l-1)})
\end{equation}

Then we use the {vertex-to-hyperedge} aggregator, namely $Agg_e$ to build a hyperedge summary from the set of related vertex features $X_v^{(l-1)}$,
\begin{equation}
        X_e^{(l)}=Agg_e^{(l)}(\{\tilde{X_v}^{(l-1)}:v\in e\})
\end{equation}

$Agg_e$ implements a reduction over the nodes. We list all $Agg_e$ we adopt in Table \ref{tab:v2eaggr}. Additionally, if available hyperedge weights $W_e$ exists, it can be applied to scale the resulting $X_e^{(l)}$. The subsequent {hyperedge-to-vertex} step collects these hyperedge summaries for each node and produces an intermediate node representation $X_v^{(l)}$ via
\begin{equation}
       \tilde{X}_v^{(l)}=Agg_v^{(l)}(\{X_e^{(l)}:e\ni v\})
\end{equation}

The choice of $Agg_v$ mirrors the design space of $Agg_e$, as shown in Table \ref{tab:e2vaggr}.

This two-stage design is motivated by the need to model true high-order relations that cannot be captured by direct node to node message passing without considering edges like graphs. Aggregating into hyperedge features allows the model to summarize group interactions and to learn hyperedge-level signals that are subsequently redistributed to nodes. This often yields richer, more expressive node representations on tasks where group membership or co-membership is informative. Furthermore, these two distinct operators together determine how high-order information is propagated within each layer and are therefore the primary sites exposed to search. {Traditional} HGNN message-passing mechanisms, however, require laborious manual design. In this work we propose to discover the best message-passing operators adaptively without manual intervention.
\begin{table}[htb!]
 \renewcommand{\arraystretch}{1.2}
\centering
\caption{{vertex-to-hyperedge} aggregation operations}
\label{tab:v2eaggr}
\begin{tabular*}{\tblwidth}{@{}LL@{}}
\toprule
$Agg_e$ & $X_e^{(l)}$\\
\midrule
sum& $H^TX_v^{(l-1)}$\\
mean& $D_e^{-1}H^TX_v^{(l-1)}$\\ 
softmax-sum& $softmax(H^T)X_v^{(l-1)}$\\
softmax-mean& $D_e^{-1}softmax(H^T)X_v^{(l-1)}$\\
\bottomrule
\end{tabular*}
\end{table}

\begin{table}[htb!]
\renewcommand{\arraystretch}{1.2}
\begin{center}
\caption{{hyperedge-to-vertex} aggregation operations}
\label{tab:e2vaggr}
\begin{tabular*}{\tblwidth}{@{}LL@{}}
\toprule
$Agg_v$ & $\tilde{X}_v^{(l)}$\\
\midrule
sum& $HX_e^{(l)}$\\
mean& $D_v^{-1}HX_e^{(l)}$\\ 
softmax-sum& $softmax(H)X_e^{(l)}$\\
softmax-mean& $D_v^{-1}softmax(H)X_e^{(l)}$\\
\bottomrule
\end{tabular*}
\end{center}
\end{table}

\subsubsection{Post-processing Layer}
% 这个写简单点吧
The post-processing layer performs normalization, activation and dropout to features $\tilde{X}_v^{(l)}$ learned by former Hyper-Interaction Module aggregators. The normalization process adjusts the input features at each layer to a consistent scale, effectively sidestepping the issue of gradient vanishing and expediting the rate at which the training converges \cite{normalization}. The normalization methods we adopt include BatchNorm \cite{batchnorm}, LayerNorm \cite{layernorm}, InstanceNorm \cite{instnorm}, GraphNorm \cite{graphnorm}, PairNorm \cite{pairnorm} and None. None indicates that no normalization operations are performed in the layer.
The activation function strengthens the model's non-linear adaptation capacity, and is a crucial element in neural network architecture design \cite{activation}. The activation function in AutoHGNN includes Relu, Relu6, Sigmoid, Tanh, Elu, LeakyRelu, SoftPlus and Identity. Identity means no activation in the layer.
Dropout is a regularization technique designed to prevent overfitting by randomly, temporarily removing a subset of neurons during training, forcing the network to learn more robust feature representations. Dropout is a common way to make the network deeper in GNNs \cite{graph-dropout}. The dropout rate we use ranges from [0,0.6] with a step size of 0.1.

Mathematically, the post-processing layer do the following operation to aggregated features
\begin{equation}
        X_v^{(l)}=Dropout(Act(Norm(\tilde{X}_v^{(l)})),dropout\_rate)
\end{equation}

\begin{table}[htb!]
 \renewcommand{\arraystretch}{1.2}
\begin{center}
\caption{Feature aggregation strategies}
\label{tab:fusion}
\begin{tabular*}{\tblwidth}{@{}CC@{}}
\toprule
Name & $Z$\\ \midrule
sum& $ \sum_{l=1}^{L-1}X_v^{(l)}$\\
mean&$ \frac{1}{L-1}\sum_{l=1}^{L-1}X_v^{(l)}$\\ 
max& $ max\{X_v^{(1)},X_v^{(2)},...,X_v^{(L-1)}\}$\\
concat& $X_v^{(1)}||X_v^{(2)}||...||X_v^{(L-1)}$\\
None& $X_v^{(L-1)}$\\
\bottomrule
\end{tabular*}
\end{center}
\end{table}
\subsubsection{Feature aggregation operator}
AutoHGNN introduces the feature aggregation mechanism to enhance the representational capacity of the network by fusing multi-scale features from different layers. Let the set of encoder outputs be denoted as $\mathcal{H}_v=\{X_v^{(1)},X_v^{(2)},...,X_v^{(L)}\}$. A feature aggregation operator $\mathcal{F}$ is applied to combine these layer-wise representations, which helps preserve both low-level and high-level semantic information and improves the model's generalization ability. The specific fusion strategies supported by $\mathcal{F}$ are detailed in Table~\ref{tab:fusion}. The aggregated representation is then passed through a lightweight decoder layer to produce the final output $Z=\mathcal{F}(\mathcal{H})$ for downstream tasks, including node classification and hypergraph classification.

\subsection{Search Module}
Before entering the search phase, {AutoHGNN first builds a weight-sharing supernet}. As shown in Fig. \ref{fig:overview} (a), the search space encompasses a set of operations denoted as $\mathcal{O}$, while the supernet comprises $\mathcal{N}$ distinct HGNN architectures that form the set $A = \{a_1, a_2, ..., a_\mathcal{N}\}$, then any two different architectures $a_i$ and $a_j$ share identical weights $\omega$ for corresponding components in their \emph{l}-th layers. This weight sharing mechanism ensures that identical components in subsequently sampled architectures inherit previously trained weights. 

Based on the supernet, the search module {combines} two submodules. The first one is a differentiable sampler, which explores and evaluates different architectures in a differentiable manner, and updates architecture $\alpha$. The second one is a supernet training module, which optimizes network weights along the sampled architectural paths.
% 他说图1(b)里没对齐就替换成下面的
% Based on the supernet constructed from the search space, the search module combains two submodules. The first one is a differentiable sampler, which explores and evaluates different architectures in a differentiable manner, and update architecture $\alpha$. The second one is a supernet training module, which optimizes network weights along the sampled architectural paths. Fig. \ref{fig:overview} (b) is the overall search module of AutoHGNN.

% \subsubsection{Weight-sharing supernet}
% Before entering the search phase, AutoHGNN first build weight-sharing a supernet. As shown in Fig. \ref{fig:overview} (b), the search space encompasses a set of operations denoted as $\mathcal{O}$, while the supernet comprises $\mathcal{N}$ distinct HGNN architectures that form the set $A = \{a_1, a_2, ..., a_\mathcal{N}\}$, then any two different architectures $a_i$ and $a_j$ share identical weights $\omega$ for corresponding components in their \emph{l}-th layers. This weight sharing mechanism ensures that identical components in subsequently sampled architectures inherit previously trained weights.
\subsubsection{Differentiable Sampler}
\label{subsubsec:sample}
To improve efficiency and stability, we employ a decoupled differentiable sampler \cite{d2nas} to sample promising architectures. This approach decouples the optimization of architecture $\alpha$ and network weights $\omega$. This allows $\alpha$ to be directly optimized via gradient descent on the validation loss. In each search epoch for a component with $j$ candidate operations, we sample a discrete selection using the Gumbel-Softmax relaxation
\begin{equation}
        \label{equ:gumbel}
s_k = \mathrm{softmax}\left(\frac{\alpha_k + g_k}{\tau}\right), \quad k = 1, \dots, j,
\end{equation}
where $g_k$ are Gumbel noise samples and $\tau$ is a temperature parameter. The operation with the highest $s_k$ is selected in each search epoch.

The architecture parameters $\alpha$ are then updated using the gradient of the validation loss
\begin{equation}
\label{equ:alpha_update}
\alpha' \leftarrow \alpha - \eta_{\alpha} \nabla_{\alpha} \mathcal{L}_{\mathrm{val}},
\end{equation}
where $\eta_{\alpha}$ is the learning rate for $\alpha$. This sampling method mitigates the bias caused by strong weight-architecture coupling in standard differentiable NAS and leads to more efficient and stable search. 

\subsubsection{Supernet Training}
This phase aims to efficiently converge the shared network weights $\omega$ towards optimal values for the promising architectures identified by the sampler. To achieve this, the update of $\omega$ and architecture parameters $\alpha$ are kept separate. Only the high-probability candidate architectures are retained, forming a focused subset of the original supernet. This strategy reduces optimization noise and allocates training resources effectively.

During each training step, an architecture is randomly sampled from the current supernet. Its inherited weights $\omega$ are updated using the training gradient on the sampled path
\begin{equation}
\label{eq:weight_update}
\omega' \leftarrow \omega - \eta \nabla_{\omega}\mathcal{L}_{\mathrm{train}},
\end{equation}
where $\eta$ is the learning rate.

Additionally, before the main search loop begins, a pre-training stage initializes $\omega$ by training on several randomly sampled architectures. This ensures a stable starting point for the subsequent optimization.

\subsection{Hypergraph Neural Network Architecture Selection}
\subsubsection{Hypergraph Stable Topological Distance}
Traditional architecture selection criteria in NAS methods, \emph{i,e.}  selecting best architectures based on validation accuracy or the maximal $\alpha$, may fail to preserve the intrinsic topology of the original data \cite{rethinking_selection}. In this section, we propose a novel Hypergraph Stable Topological Distance (HyperSTD) to effectively select top-performing HGNN architectures based on measuring the consistency of learned representations with original higher-order structures.

To begin with, motivated by the architecture selection strategy of Auto-HeG \cite{auto_heg}, we introduce Hypergraph Topology Affinity Matrices (HTAM) that quantify how well predicted labels preserve hypergraph connectivity. The original HTAM is defined in Eq. \eqref{equ:htam}.
\begin{equation}
        \label{equ:htam}
        \begin{split}
                &S=(Y^{\top}\Theta Y)\oslash(Y^{\top}\Theta E + \varepsilon )\\
                &\hat{S}=(\hat{Y}^{\top}\Theta\hat{Y}) \oslash(\hat{Y}^{\top}\Theta E+\varepsilon)
        \end{split}
\end{equation}
where $\Theta=HW_eD_{e}^{-1}H^{\top}$ is the hypergraph adjacency matrix, with $Y, \hat{Y}\in\mathbb{R}^{n\times C}$ are one-hot class encodings for ground-truth and predicted labels, $E\in\mathbb{R}^{n\times C}$ is an all-ones matrix, and $\oslash$ denotes elementwise division. Typically, all hyperedges have the same weights, in this case $W_e$ is the identity matrix $I$, so $\Theta=HD_{e}^{-1}H^{\top}$. We also add the small constant $\varepsilon > 0$ in the denominators to prevent division-by-zero or extreme amplification when certain classes are rare within particular hyperedges.

Second, rather than relying on the raw distance metric in Auto-HeG \cite{auto_heg}, we define the HyperSTD to create a common ground for comparing across various datasets and architectures, as shown in Eq. \eqref{equ:HyperSTD}.
\begin{equation}
        \label{equ:HyperSTD}
        D_{\text{HyperSTD}}=\frac{||\hat{S}-S||_F}{||S||_F}.
\end{equation}

{The relative Frobenius distance in Eq. \eqref{equ:HyperSTD} is scale-invariant and bounded because $S \in [0,1]$. However, it can be sensitive when $\|S\|_F$ is extremely small or when $S$ is nearly singular in practice. In our case, $S$ is well-conditioned for the datasets used (see Table \ref{tab:dataset}). For future work, we may investigate more robust topology-aware distances, such as cosine similarity after low-rank approximation or Hypergraph aware variants of graph edit distance.}

A small HyperSTD indicates that the predicted label affinities closely align with the original hypergraph structure, providing a fair and interpretable criterion for comparing architectures. Thus, HyperSTD serves as a principled and topology-aware metric for architecture selection.

\subsubsection{Two-stage Architecture Selection with HyperSTD}
\label{subsec:arch_selection_with_hyperstd}
The architecture selection in AutoHGNN follows a two-stage process designed to progressively refine the candidate pool, first by filtering within each search round and then by selecting globally from all accumulated promising architectures. This selection process is illustrated in Fig.~\ref{fig:overview}(c).

The first stage selection is the promising architecture screening. It is performed after each round of differentiable sampling while a set of candidate architectures is sampled. From this set, we compute the HyperSTD value for each candidate using Eq.~\eqref{equ:HyperSTD}. We then select the top-$K$ architectures with the smallest HyperSTD values as local promising candidates. These architectures are considered promising and are added to a promising candidate pool named $\mathcal{C}$. This stage operates within each search iteration, ensuring that only structurally faithful architectures from the current round are retained for further consideration.

We perform a second-stage selection since the supernet weights are updated through training after each round of sampling, the quality of sampled architectures gradually improves over subsequent rounds. This stage is named global architecture selection. In this stage, we select the top-$K$ architectures with the lowest HyperSTD scores from $\mathcal{C}$. Each architecture in these top-$K$ candidates is then fully retrained on the training set and evaluated on the validation set under the standard training protocol. The final architecture is chosen as the one achieving the highest validation accuracy from the top-$K$ candidates. This step first filters by structural consistency and then by task accuracy to ensure that the chosen model is both topologically faithful and high-performing.

\subsection{Pseudocode of AutoHGNN}
Algorithm \ref{alg:overall} summarizes the comprehensive search and selection procedure of AutoHGNN. The algorithm integrates the differentiable search strategy and the topology-aware selection via HyperSTD. This structured pipeline ensures an efficient and robust search for optimal hypergraph neural architectures.

\begin{breakablealgorithm}
        \vspace{2pt}
\caption{AutoHGNN}
\label{alg:overall}
\begin{algorithmic}[1]
\renewcommand{\algorithmicrequire}{\textbf{Input:}}
\renewcommand{\algorithmicensure}{\textbf{Output:}}

\REQUIRE HGNN search space $S^*$, training set $G_{train}$, validation set $G_{val}$, temperature $\tau$, total epochs $j$, search epochs $k$, sample size $s$, HyperSTD return size $K$
\ENSURE Optimal HGNN architecture $a^*$
\STATE \textbf{// Build and initialize weight-sharing supernet}
\STATE $\mathcal{N} \gets \text{build\_supernet}(S^*)$ // {construct supernet from search space $S^*$}
\STATE $\mathcal{C} \gets \varnothing$ // promising architectures

\FOR{$i=1$ to $s$}
\STATE Sample $a$ from $\mathcal{N}$
\STATE Train $a$ on $G_{train}$ to initialize weight $\omega$ // pre-training
\ENDFOR
\STATE \textbf{// Differentiable Sampler}
\FOR{$iter=1$ to $j$}
\STATE $\mathcal{H} \gets \varnothing$ 
\FOR{$epoch=1$ to $k$}
\STATE $\alpha \gets \text{GumbelSoftmax}(\alpha, \tau)$ (Eq.~\eqref{equ:gumbel})
\STATE Sample $a$ from $\mathcal{N}$ using maximum $\alpha$
\STATE Update $\alpha$ on $G_{val}$ via Eq.~\eqref{equ:alpha_update}
\STATE $b \gets \arg\max_{\alpha} \mathcal{N}$
\STATE $\mathcal{H} \gets \mathcal{H} \cup \{b\}$
\ENDFOR

\STATE \textbf{// Stage 1 architecture selection}
\STATE Compute HyperSTD for $\mathcal{H}$ via Eq.~\eqref{equ:HyperSTD}
\STATE $\mathcal{R} \gets \text{minimum}_K(\mathcal{C})$  // HyperSTD
\STATE $\mathcal{C} \gets \mathcal{C} \cup \mathcal{R}$
\STATE $\mathcal{N} \gets \mathcal{H}$
\STATE \textbf{// Supernet Training}
\FOR{$epoch=1$ to $s$}
\STATE Sample $a$ from $\mathcal{N}$
\STATE Train $a$ on $G_{train}$ via Eq.~\eqref{eq:weight_update} // supernet training
\ENDFOR
\ENDFOR
\STATE \textbf{// Stage 2 architecture selection}
\STATE Compute HyperSTD for $\mathcal{C}$ via Eq.~\eqref{equ:HyperSTD}
\STATE $\mathcal{F} \gets \text{minimum}_K(\mathcal{C})$ // final candidates
\STATE Re-evaluate $\mathcal{F}$ on $G_{val}$
\RETURN Best architecture $a^*$ from $\mathcal{F}$
\end{algorithmic}
\end{breakablealgorithm}

%这段要写？The selected architecture is subsequently retrained from scratch under the full training regime and evaluated on the held-out test set.
\section{Experiment}
\subsection{Datasets}

To assess the performance of AutoHGNN, we conduct experiments utilizing three widely recognized graph benchmarks and two hypergraph datasets. For graph datasets, Pubmed is a well-established citation network commonly employed in graph-based research. Computers is an Amazon co-purchase network from the e-commerce domain. Physics is a Microsoft Academic graph capturing co-authorship relationships among researchers. For hypergraph datasets, DBLP and Cora Co-authorship(Cora-CA) are co-authorship hypergraphs, and we get them in THU-DeepHypergraph toolbox proposed in HGNN+ \cite{hgnnp}. Details of the datasets are in Table \ref{tab:dataset}.

% Since graph is a special case of hypergraph, we utilize three open graph benchmark datasets that are popular for GNAS research.
\begin{table*}[htb!]
    \centering
    \caption{Details of datasets}
    \begin{tabular*}{\tblwidth}{@{}LLLLL@{}}
    \toprule
        Dataset Name & Nodes & Edges/Hyperedges & Feature dims & Classes \\
        \midrule
        Pubmed & 18,333 & 81,894 & 6,805 & 15 \\
        Computers & 13,381 & 245,778 & 767 & 10 \\
        Physics & 34,493 & 247,962 & 8,415 & 5 \\
        Cora\_CA & 2,708 & 1,072 & 1,433 & 7 \\ 
        DBLP & 41,302 & 22,363 & 1,425 & 6  \\
        \bottomrule
    \end{tabular*}
    \label{tab:dataset}
\end{table*}

\subsection{Baselines and Metrics}
When evaluating the PubMed, Physics, and Computers graph datasets, our proposed method will be benchmarked against  hand-crafted GNN architectures, including GAT \cite{GAT}, GraphSAGE \cite{graphsage}, SGC \cite{SGC}, GraphConv \cite{graphconv} and GATv2 \cite{gatv2}, non-differentiable NAS approaches including GraphNAS \cite{graphnas}, DeepGNAS \cite{deepgnas} and AutoGNAS\cite{auto_gnas} as well as differentiable NAS methods including DARTS \cite{darts}, DSS \cite{one_shot_gnas} and D2GNAS \cite{d2gnas}.
For the Cora-CA and DBLP hypergraph datasets, except for the above-mentioned GNN models, we compare our method with hypergraph methods including HGNN \cite{hgnn}, HGNN+ \cite{hgnnp}, HyperGCN \cite{hypergcn}, UniGNN \cite{unignn}, UniG-Encoder \cite{unig_encoder} and its variant UniG-EncoderII with hops set to two and three. UniGNN includes UniGCN, UniGAT, UniSAGE and UniGIN. Since prior NAS techniques were not developed for hypergraph neural networks rather than hypergraph structures, we do not evaluate the performance using NAS methods on these datasets.
For the node classification task, we use accuracy as a metric to evaluate model performance. Average test accuracy serves as the metric for assessing the generalization capability of all methods. The promising configuration of AutoHGNN search parameters is evaluated based on the validation accuracy obtained in the search parameter sensitivity analysis experiment. In the efficiency comparison experiment, we use the average search time per epoch (STPE) as the time efficiency metric.

\subsection{Experimental Setup}
\textbf{Overview} Graph datasets will be transformed into hypergraphs utilizing the k-hop methodology described in HGNN+ \cite{hgnnp}. To ensure compatibility with the message-passing mechanism for hypergraphs within the search space while preserving as much of the original graph's structural information as possible, we have set k equal to 1. We {conduct} our experiments on a server with 1 NVIDIA Geforce RTX 2080 Ti GPU with 11GB memory.

%but with a batch normalization within two layers，审稿人问就解释一下说不加的话效果过差
\textbf{Implementation details} All manual GNN architectures feature two layers with a hidden dimension of 128. For GAT and GATv2 models, the attention mechanism has two heads. All other parameter values remained at their defaults as provided in the PyTorch Geometric\footnote{\url{https://pytorch-geometric.readthedocs.io/en/}} implementation. For manual HGNNs, we adopt a similar structural blueprint but with a Batch Normalization within two layers. The Adam optimizer served as our choice for all manual approaches, configured with a learning rate of 0.005 and a weight decay of 1e-4. As for the graph NAS methods, we {used the exact same hyperparameter settings} that were employed in their papers' experimental section. All experiments use the standard cross-entropy loss function.

For AutoHGNN, we configure the network depth $L$ to 2. The total search iterations $j$ is 2, with search rounds $k$ established at 500, alongside s=100. The temperature parameter $\tau$ is set to 0.2. Throughout the search process, the Adam optimizer is used for both $\alpha$ and $\omega$ parameters with a consistent learning rate of 0.01. For the $\alpha$ optimization, the weight decay is 0.005, whereas in the $\omega$ optimization it is 0.0001. We let HyperSTD return $K$=5 candidate architectures. In the re-evaluation phase, we use tree-structured parzen estimators (TPE) \cite{tpe} in the Python package HyperOpt\footnote{\url{http://hyperop.github.io/hyperopt/}} to determine the optimal hyperparameters of each candidate, followed by conducting 10 rounds of testing on the selected best architecture. Both the re-evaluation and testing phases have 100 training epochs. The specific hyperparameter groups are as follows:
\begin{itemize}
        \item Learning rate (lr): [0.1, 0.01, 5e-3, 1e-3, 1e-4]
        \item Weight decay (decay): [0, 1e-5, 5e-5, 1e-4, 5e-4, 1e-3]
        \item Hidden dimensions (hidden\_dim): [64, 128, 256, 384]
\end{itemize}

\subsection{Results}
We calculated the average and standard deviation of the test accuracy of the optimal architecture selected by the testing phase. The results of the node classification task on graph and hypergraph datasets are listed in Table \ref{tab:result_graph} and Table \ref{tab:result_hypergraph}, respectively.

\textbf{Result analysis on graph datasets}. On graph benchmarks, the proposed approach matches or surpasses recent manual GNN and graph NAS baselines while maintaining low search cost, as is suggested in Table \ref{tab:result_graph}. For instance, AutoHGNN achieves a test accuracy of 95.53\% on the Physics dataset, outperforming the best baseline, D2GNAS, by approximately 0.78 percentage points. This gain can be attributed to our hypergraph-formulated search space which, even when applied to graphs converted to 1-hop hypergraphs, encourages richer aggregation beyond immediate pairwise neighbors. Notably, the efficiency of our decoupled search strategy results in a per-epoch search time nearly an order of magnitude lower than reinforcement learning- or evolutionary-based NAS methods, making the framework both more accurate and practical.

\textbf{Result analysis on hypergraph datasets}. Across hypergraph benchmarks, our method yields consistently stronger performance than a broad set of GNN and HGNN baselines. As shown in Table \ref{tab:result_hypergraph}, AutoHGNN increases at least 1.07\% accuracy on Cora-CA and 2.88\% on DBLP. This is a clear improvement over the best manually designed HGNN and other strong baselines. The improvement is more pronounced here due to the exact match between our Hyper-Interaction Module search space and the intrinsic higher-order structure of hypergraphs. The compact space allows efficient discovery of effective aggregation pairs. Furthermore, the final performance is bolstered by the HyperSTD selection, which filters for architectures that preserve the original hypergraph’s structural affinities, leading to models with low performance variance. This demonstrates the necessity of co-designing the search space and a topology-aware selection criterion for hypergraph learning.

\begin{table*}[htb!]
\centering
\caption{Result on graph datasets. Bold and underlined texts indicate best and second-best models respectively. STPE indicates Search Time Per Epoch for NAS methods.}
\label{tab:result_graph}
\begin{tabular*}{\tblwidth}{@{}LLLLLLL@{}}
\toprule
Method & \multicolumn{2}{c}{Pubmed} & \multicolumn{2}{c}{Computers} & \multicolumn{2}{c}{Physics} \\ \cline{2-7}
 & Acc(\%) & STPE(s) & Acc(\%) & STPE(s) & Acc(\%) & STPE(s) \\ \midrule
GAT &78.85$\pm$0.55  &\, - &75.89$\pm$0.66  &\, - &92.19$\pm$0.66  &\, - \\
GraphSAGE &80.90$\pm$0.19  &\, - &78.34$\pm$0.48  &\, - &92.90$\pm$0.56  &\, - \\
SGC &78.86$\pm$0.27  &\, - &81.87$\pm$0.56  &\, - &94.49$\pm$0.22  &\, - \\
GraphConv &75.88$\pm$0.23 &\, - &76.03$\pm$0.30 &\, - &89.04$\pm$1.55 &\, -\\
GATv2 &79.82$\pm$0.41 &\, - &76.19$\pm$0.26 &\, - &93.00$\pm$0.70 &\, - \\
GraphNAS &81.16$\pm$0.21  &\, 17.946  &85.50$\pm$0.10  &\, 42.134  &93.51$\pm$0.08  &\, 32.811  \\
DeepGNAS &80.17$\pm$0.12  &\, 18.265  &85.83$\pm$0.14  &\, 42.910  &93.95$\pm$0.17  &\, 43.450  \\
AutoGNAS &81.86$\pm$0.34  &\, 12.044  &\underline{85.99$\pm$0.25}  &\, 38.525  &94.49$\pm$0.09  &\, 42.010  \\
DARTS &81.75$\pm$0.32  &\, 0.180  &85.43$\pm$0.31  &\, 0.447  &93.84$\pm$0.09  &\, 0.537  \\
DSS &81.89$\pm$0.13  &\, 0.189  &85.80$\pm$0.21  &\, 0.594  &93.59$\pm$0.14  &\, 0.543  \\
D2GNAS &\underline{82.18$\pm$0.41}  &\, 0.183  &85.95$\pm$0.25  &\, 0.422  &\underline{94.75$\pm$0.18}  &\, 0.501  \\
\textbf{AutoHGNN} & \textbf{83.17$\pm$0.23}  &\, \textbf{0.137}  &\textbf{87.02$\pm$0.17}  &\, \textbf{0.110}  &\textbf{95.53$\pm$0.04}  &\, \textbf{0.218} \\ \bottomrule
\end{tabular*}
\end{table*}

\begin{table}[htb!]
\centering
\caption{Result on hypergraph datasets. Bold and underlined texts indicate best and second-best models respectively.}
\label{tab:result_hypergraph}
\begin{tabular*}{\tblwidth}{@{}LLLLL@{}}
\toprule
Method & \multicolumn{2}{c}{Cora\_CA} & \multicolumn{2}{c}{DBLP} \\ \cline{2-5} 
 & Mean (\%) & Std (\%) & Mean (\%) & Std (\%) \\ \midrule
GAT & 66.95 & 1.00 & 83.06 & 0.48 \\
GraphSAGE & 72.38 & 0.95 & 84.69 & 0.09 \\
SGC & 66.56 & 0.47& 83.05 & 0.32 \\
GraphConv & 70.71 & 0.40 & 83.17 & 0.37 \\
GATv2 & 67.01 & 0.77 & 83.68 & 0.70 \\
HGNN & 67.55 & 0.28 & 77.31 & 1.34 \\
HGNN+ & 67.17 & 0.38 & 75.89 & 0.94 \\
HyperGCN & 69.06 &1.32 & 73.57 & 1.62 \\
UniGCN & 66.14 & 0.28 & 87.02 & 0.07 \\
UniGAT & 65.72 & 0.28 & 86.40 & 0.09 \\
UniSAGE & 72.25 & 0.30 & \underline{87.52} & \underline{0.06} \\
UniGIN & \underline{72.69} & \underline{0.86} & 87.38 & 0.10 \\
UniG-Encoder & 71.07 &0.52 &83.71 & 0.33\\
UniG-EncoderII (2-hop) & 70.58 &1.26 &83.04 &0.21\\
UniG-EncoderII (3-hop) & 72.38 &0.63 &86.03 &0.09\\
\textbf{AutoHGNN} & \textbf{73.76} & \textbf{0.41} & \textbf{88.91} & \textbf{0.07} \\ \bottomrule
\end{tabular*}
\end{table}

\subsection{Sensitivity Analysis}
\begin{figure*}[htb!]
\centering
\subfloat{\includegraphics[width=70mm]{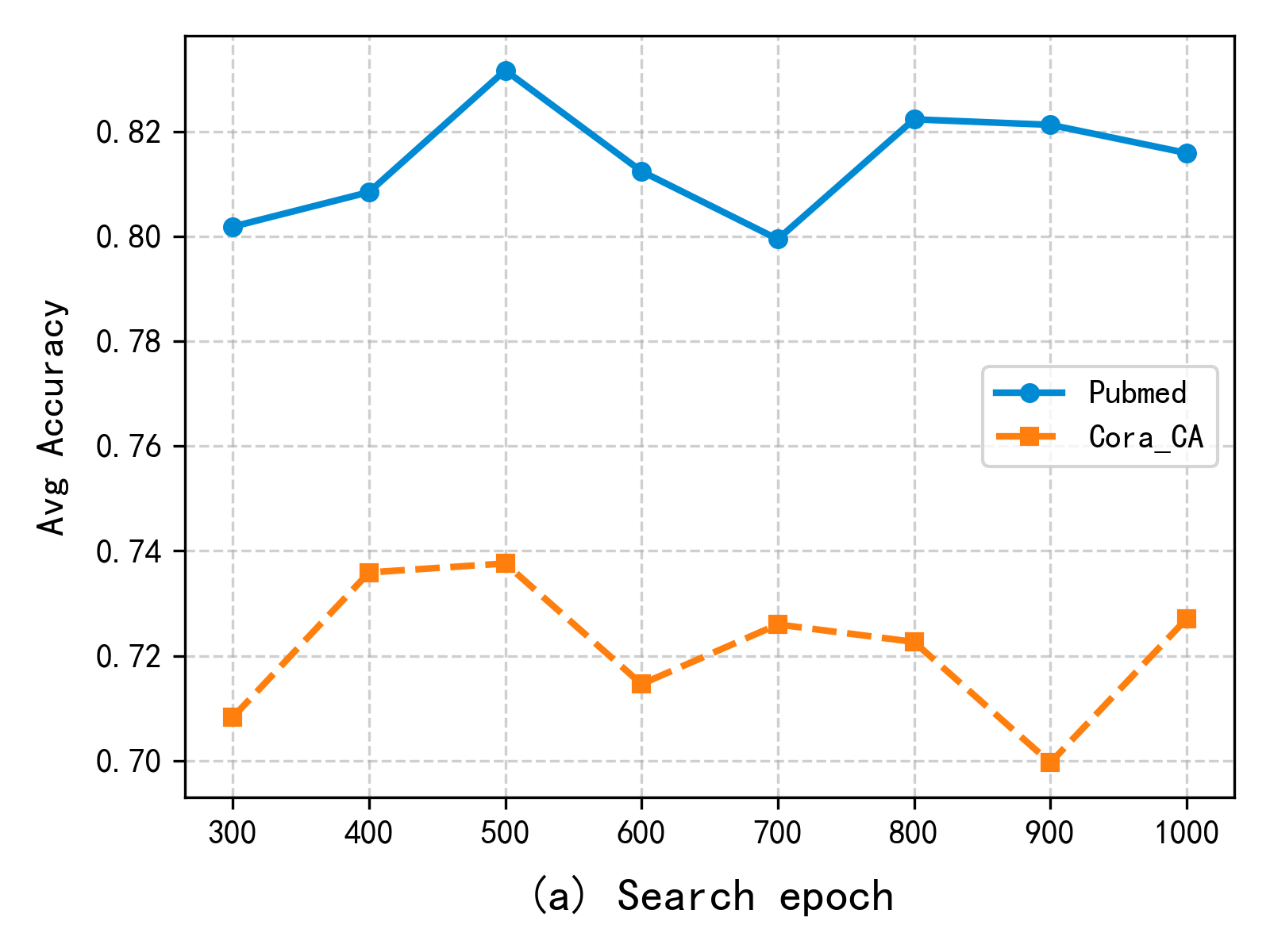}} % search epoch
\subfloat{\includegraphics[width=70mm]{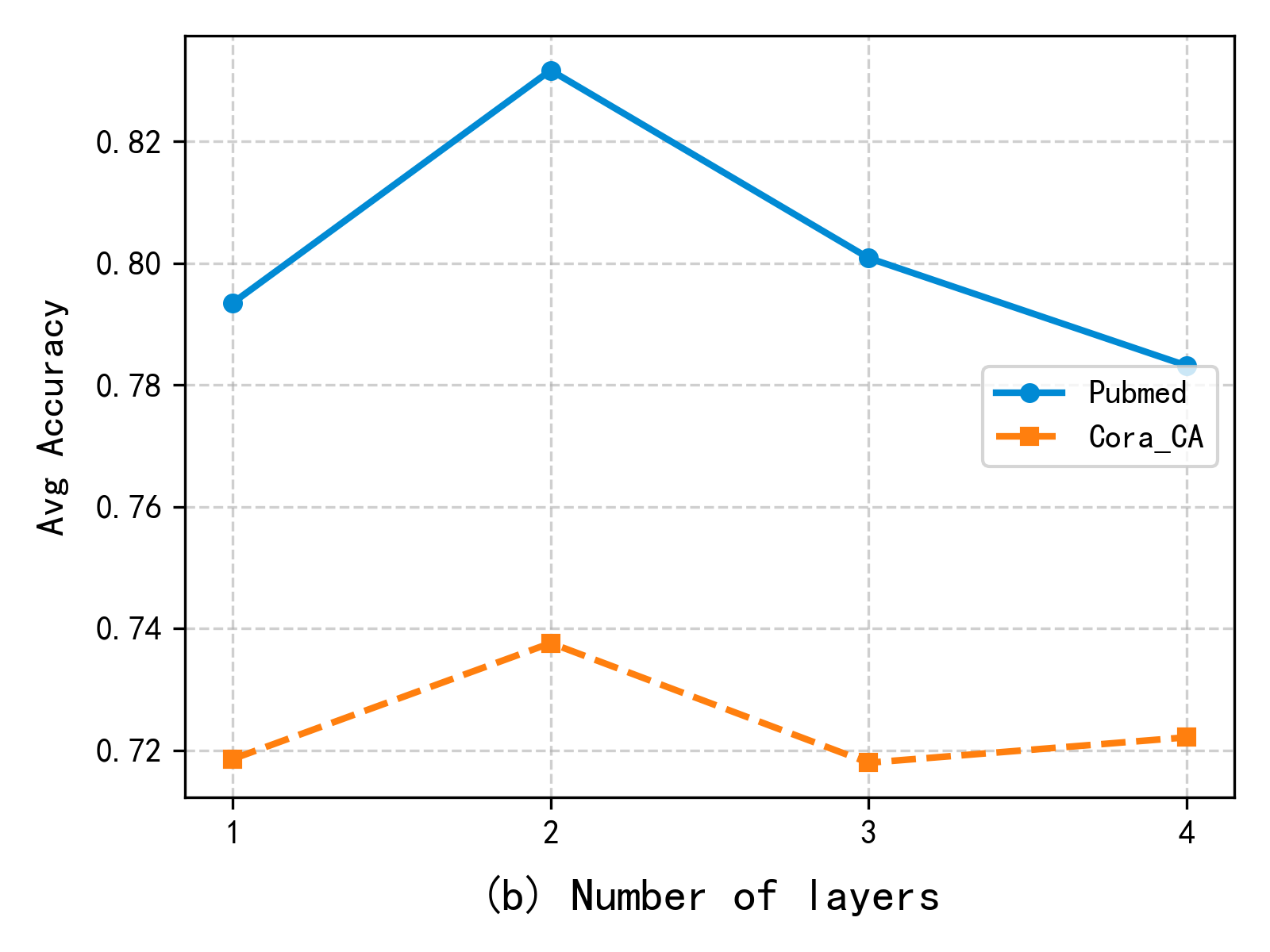}}\hfill % layers
\subfloat{\includegraphics[width=70mm]{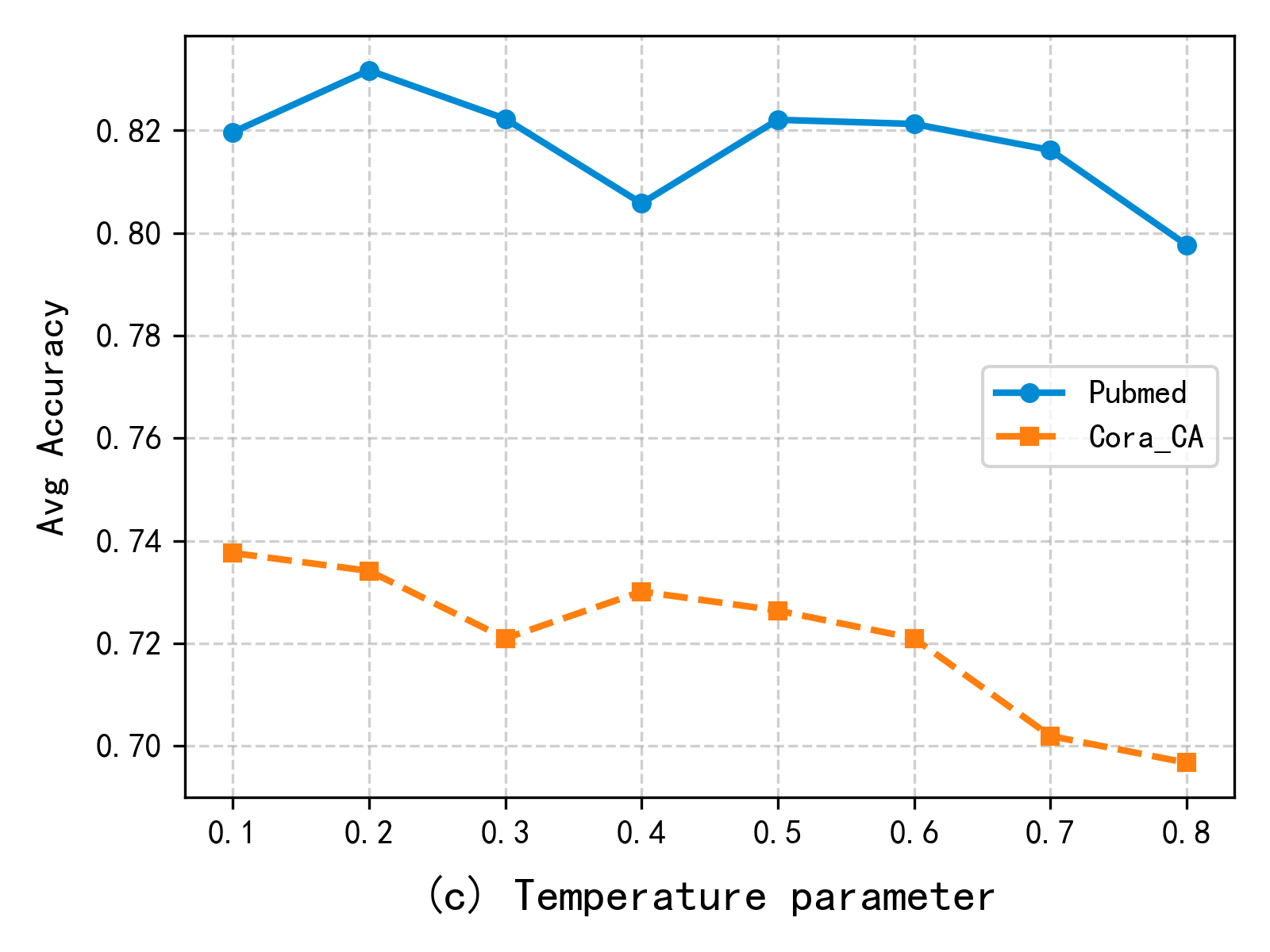}} % temprature
\subfloat{\includegraphics[width=70mm]{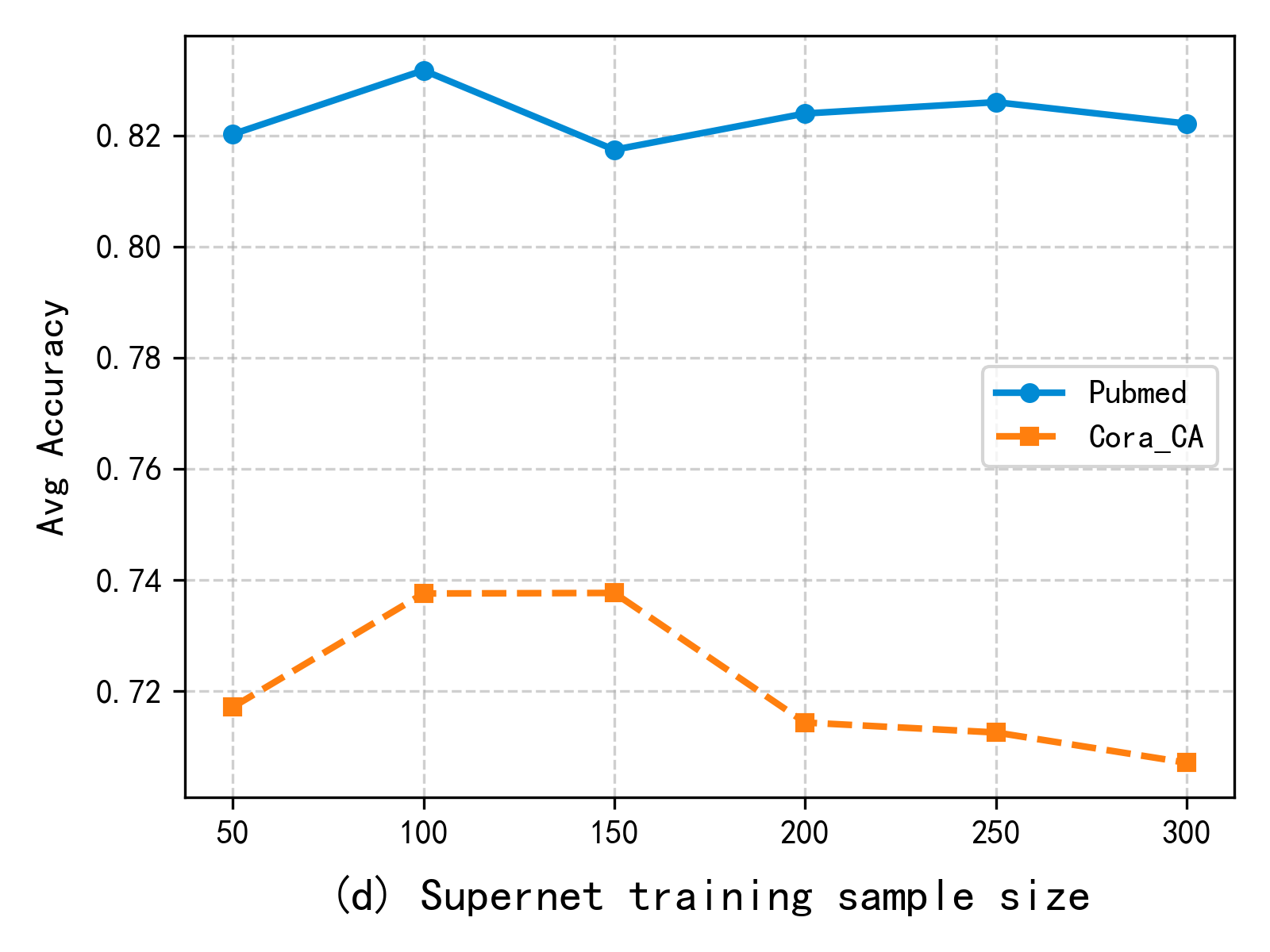}} % sample size
\caption{Hyperparameter sensitivity analysis of AutoHGNN Hyperparameter sensitivity analysis of AutoHGNN on (a) search epochs, (b) network depth, (c) temperature, and (d) training sample size. The X axis is the parameter value and Y axis is the average accuracy of 10 experiments on best architecture found by AutoHGNN.}
\label{fig:sensitivity}
\end{figure*}

We conduct parameter sensitivity analysis experiments to determine optimal search parameters for AutoHGNN by evaluating their impact on validation performance. For the Pubmed graph dataset alongside the Cora\_CA hypergraph dataset, we evaluate our approach by examining the average validation accuracy achieved by the five most promising HGNN architectures discovered through our method. This investigation focuses on four critical search parameters: differentiable search epoch $k$, temperature coefficient $\tau$, supernet training sample size $s$, network layer $L$. To explore how various parameters affect the overall performance of AutoHGNN, we turned to the control variable approach. Firstly, we discuss the differentiable search epoch parameter while keeping the supernet training sample size of 100, the temperature coefficient of 0.1, and just 2 network layers. Once we find the best parameters, we lock them in place and move on to examine the subsequent parameters in line. All results are shown in Fig. \ref{fig:sensitivity}.

The larger the search epoch, the more expansive the architecture whose weight $\alpha$ is updated becomes. This broader landscape allows AutoHGNN to identify a promising HGNN design. Nevertheless, when it becomes excessively large, it creates a massive search space within the supernet, which makes it more challenging for the differentiable search to pinpoint viable HGNN architectures in the supernet. Fig. \ref{fig:sensitivity} (a) shows that the best epoch is 500. Deeper layers can capture longer-range dependencies and complex patterns, but they will lead to over-smoothing, vanishing gradients, noise accumulation, or over-fitting, diluting information. Too few layers may cause insufficient expressiveness and cannot integrate enough context. Fig. \ref{fig:sensitivity} (b) shows that the layer number of two is the best on these data sets. The temperature parameter $\tau$ essentially dictates the balance between exploration and exploitation in decoupled differentiable search. When it is set low, the search process tends to get stuck on promising HGNN architectures it has already uncovered. Conversely, a higher $\tau$ value broadens the search horizons, enabling the algorithm to discover new potential HGNN architectures. According to Fig. \ref{fig:sensitivity} (c), we select 0.2 as the best temperature. When it comes to Supernet training, updating the $\omega$ parameter is absolutely crucial for optimal architecture gradients. On top of that, it gives us a solid starting point for initializing $\omega$. However, too large training sample size might just increase the probability of overfitting. Fig. \ref{fig:sensitivity} (d) indicates that 100 is a suitable sample size.

\subsection{Ablation Study}
We conduct ablation studies on different datasets to confirm the contribution of each core component in AutoHGNN. The ablation results in Table \ref{tab:ablation} show that the complete AutoHGNN achieves the best performance across all five datasets.

The variant with fixed, identical aggregation operators (mean+mean) across both message-passing stages shows a clear performance drop across all datasets, with the most significant degradation observed on the hypergraph datasets (Cora-CA and DBLP). This confirms that a flexible search space capable of adaptively selecting different aggregation operators for the vertex-to-hyperedge and hyperedge-to-vertex stages is essential for capturing the complex higher-order interactions inherent in hypergraph data.

Removing the feature aggregation mechanism (w/o fusion) leads to degraded performance on most datasets, with an average accuracy drop of approximately 1–3 percentage points. This underscores the importance of fusing multi-scale features from different layers, which helps preserve both low-level and high-level semantic information and improves the model's generalization ability. The effect is particularly noticeable on the larger datasets (Computers and Physics), where multi-scale representations are more critical for capturing complex patterns.

The variant that selects architectures without the HyperSTD metric (w/o HyperSTD) exhibits performance instability and a consistent drop in final accuracy, especially on hypergraph datasets. This demonstrates that HyperSTD effectively filters architectures that preserve the original hypergraph's structural affinities, leading to more robust and topologically faithful models. The higher standard deviation observed in the w/o HyperSTD variant further indicates that the selection process becomes less stable without this topology-aware criterion.

Overall, the ablation study confirms that the compact search space with flexible message aggregation operators, the feature aggregation mechanism, and the HyperSTD-based selection criterion are all integral components contributing to the robustness, stability, and superior performance of the proposed AutoHGNN framework. The consistent improvements across both graph and hypergraph benchmarks highlight the generality and effectiveness of our design choices.

Therefore, the ablation study confirms that the compact search space with flexible message aggregation operators, the feature aggregation mechanism, and the HyperSTD-based selection criterion are all integral components contributing to the robustness, stability, and superior performance of the proposed AutoHGNN framework.

\section{Conclusion}
This paper presents a novel neural architecture search framework for hypergraph neural networks, namely AutoHGNN. It introduces a compact search space that explicitly models vertex-hyperedge-vertex message propagation. Furthermore, the proposed Hypergraph Stable Topological Distance (HyperSTD) metric selects architectures that preserve original hypergraph structural affinities after the decoupled differentiable search strategy. The whole framework provides a practical approach for automatically discovering effective models that exploit higher-order relations in structured data. Experiments on graph and hypergraph benchmark datasets show that AutoHGNN outperforms both manual and automated baselines in accuracy with lower search cost.

{\textbf{Implications, limitations and future work.} AutoHGNN provides a practical solution for automating hypergraph model design. However, the current search space is limited to two-stage message passing, and HyperSTD can be sensitive under extreme circumstances. Future work includes: (1) extending the search space with attention mechanisms and residual connections; (2) exploring more robust topology-aware metrics beyond the Frobenius norm, such as cosine similarity after low-rank approximation; and (3) applying AutoHGNN to large-scale dynamic hypergraphs and semi-supervised scenarios.}

\begin{table*}[htb!]
\centering
\caption{Ablation study results. Bold texts indicate the best model.}
\label{tab:ablation}
\begin{tabular}{@{}llllll@{}}
\toprule
Variant & Pubmed & Computers & Physics & Cora\_CA & DBLP \\ \midrule
AutoHGNN(mean+mean) &81.95$\pm$0.12 &84.92$\pm$0.09  &93.82$\pm$0.08  &67.26$\pm$0.21  &77.02$\pm$0.11 \\
AutoHGNN(w/o fusion) &79.98$\pm$0.65  &84.39$\pm$0.37  &93.97$\pm$0.16  &66.18$\pm$0.12  &86.47$\pm$0.20  \\
AutoHGNN(w/o HyperSTD) &82.01$\pm$0.11  &85.10$\pm$0.09  &94.85$\pm$0.03  &72.65$\pm$0.44  &88.09$\pm$0.09  \\
AutoHGNN & \textbf{83.17$\pm$0.23} & \textbf{87.02$\pm$0.17} & \textbf{95.53$\pm$0.04} & \textbf{73.76$\pm$0.41} & \textbf{88.91$\pm$0.07} \\ \bottomrule
\end{tabular}
\end{table*}

\printcredits

\bibliographystyle{model1-num-names}
% \bibliographystyle{cas-model2-names}

\bibliography{references}



\end{document}